\pdfoutput=1

\documentclass{article}

\usepackage[latin1]{inputenc}% erm\"oglich die direkte Eingabe der Umlaute 
\usepackage[T1]{fontenc} % das Trennen der Umlaute
\usepackage{arxiv}
\usepackage{natbib}
\usepackage{graphicx}
\usepackage{float}
\title{How Much Can We See? A Note on Quantifying Explainability of Machine Learning Models}
\author{Gero Szepannek \\
	Stralsund University of Applied Sciences \\
	Zur Schwedenschanze 15 \\
	18435 Stralsund \\
	Germany \\
	}

\begin{document}

\maketitle

\begin{abstract}
One of the most popular approaches to understanding feature effects of modern black box machine learning models are partial dependence plots (PDP). These plots are easy to understand but only able to visualize low order dependencies. The paper is about the question \emph{''How much can we see?''}: A framework is developed to quantify the \emph{explainability} of arbitrary machine learning models, i.e. up to what degree the visualization as given by a PDP is able to explain the predictions of the model. The result allows for a judgement whether an attempt to explain a black box model is sufficient or not.\end{abstract}

\section{Introduction}\label{intro}
In the recent past a considerable number of auto machine learning frameworks such as \texttt{H2O}, \texttt{auto-sklearn} \citep{auto-sklearn} or \texttt{mlr3} \citep{mlr} have been developed and made publicly available and thus simplify creation of complex machine learning models. On the other hand, advances in hardware technology allow these models to get more and more complex with huge numbers of parameters such as deep learning models \citep[cf. e.g.][]{lecun15}. Properly parameterized modern ML algorithms are often of superior predictive accuracy. 

The popularity of modern ML algorithms is based on the fact that they are very flexible with regard to    
to detection of complex nonlinear high dimensional multivariate dependencies without the need for an explicit specification of the type of the functional relationship of the dependence. As a consequence the resulting models are often called to be of black box nature which has led to an increasing need of tools for their interpretation.

Depending on the context, there are different requirements to explainability \citep[cf. e.g.][]{dalex, szepannek19} given by different targets of explanation
such as explanations of predictions for individual observations \citep{ribeiro16,strumbelj14, lundberg17,biecek18}, importance of features \citep{breiman01,casalicchio18} and feature effects \citep{friedman01,apley16,goldstein15}. 

This paper concentrates on the latter: feature effects do investigate the dependency of the  predictions by a model on one (or several) predictors. \cite{molnar19} work out that superior performance comes along with the ability to model nonlinear high oder dependencies which are naturally hard to understand for humans. As a remedy, criteria are developed in order to quantify the interpretability of a model and in consequence allow for multi-objective optimization of the model selection process with respect to both: predictive performance and interpretability.  

The approach in this paper is somewhat different: Starting with any model (which is often the best one in terms of predcitive accuracy) one of the most popular approaches to understanding of feature effects are partial dependence plots (PDP) which are introduced in Section~\ref{pdp}. Partial dependence plots are easy to understand but only able to visualize low order dependencies. The question that is asked in this paper is \emph{''How much can we see?''}: In Section~\ref{explainability} a framework is developed to quantify the \emph{explainability} of a model, i.e. up to what degree the visualization as given by a PDP is able to explain a model. This allows us to judge whether an attempt to explain the predictions of a model is sufficient or not. In Section~\ref{examples} the approach is demonstrated on two examples uing both artificial as well as real-world data and finally, a summary and an outlook are given in Section~\ref{summary}.

\section{Partial Dependence}\label{pdp}
Partial depencence plots \citep[PDP][]{friedman01} are a model-agnostic approach in order to understand feature effects and are applicable to arbitrary models, here denoted by $\hat{f}(x)$. The vector of predictor variables $x = (x_s, x_c)$ is further subdivided into two subsets: $x_s$ and $x_c$. The partial dependence function is given by
\begin{equation}\label{pd}
PD_s(X) = PD_s(X_s) = \int \hat{f}(X_s, X_c) dP(X_c), 
\end{equation}
i.e. it computes the average prediction given the variable subset $X_s$ takes the values $x_s$. In practise, the partial dependence curve is estimated by
\begin{equation}\label{pdemp}
\widehat{PD}_s(x) = \widehat{PD}_s(x_s) =\frac{1}{n} \sum_{i=1}^n \hat{f}(x_s, x_{ic}). 
\end{equation}

Note that for $X_s = X$ the partial dependence function $PD_s(x)$ corresponds to $\hat{f}(x)$ and
in the extreme, for the variable subset $s = \emptyset$, i.e. $X_c = X$, this will end up in: 
\begin{equation}\label{avpd}
PD_{\emptyset}(X) = PD_{\emptyset} = \int \hat{f}(X) dP(X), 
\end{equation}
which is independent of $x$ and corresponds to the constant average prediction of the model estimated by: 
\begin{equation}\label{avpdemp}
\widehat{PD}_{\emptyset}(x) = \widehat{PD}_{\emptyset} = \frac{1}{n} \sum_{i=1}^n \hat{f}(x_{i}). 
\end{equation}

\section{Explainability}\label{explainability}

\begin{figure}\label{fig:sim1}
\begin{center}
	\includegraphics[width=0.9\textwidth]{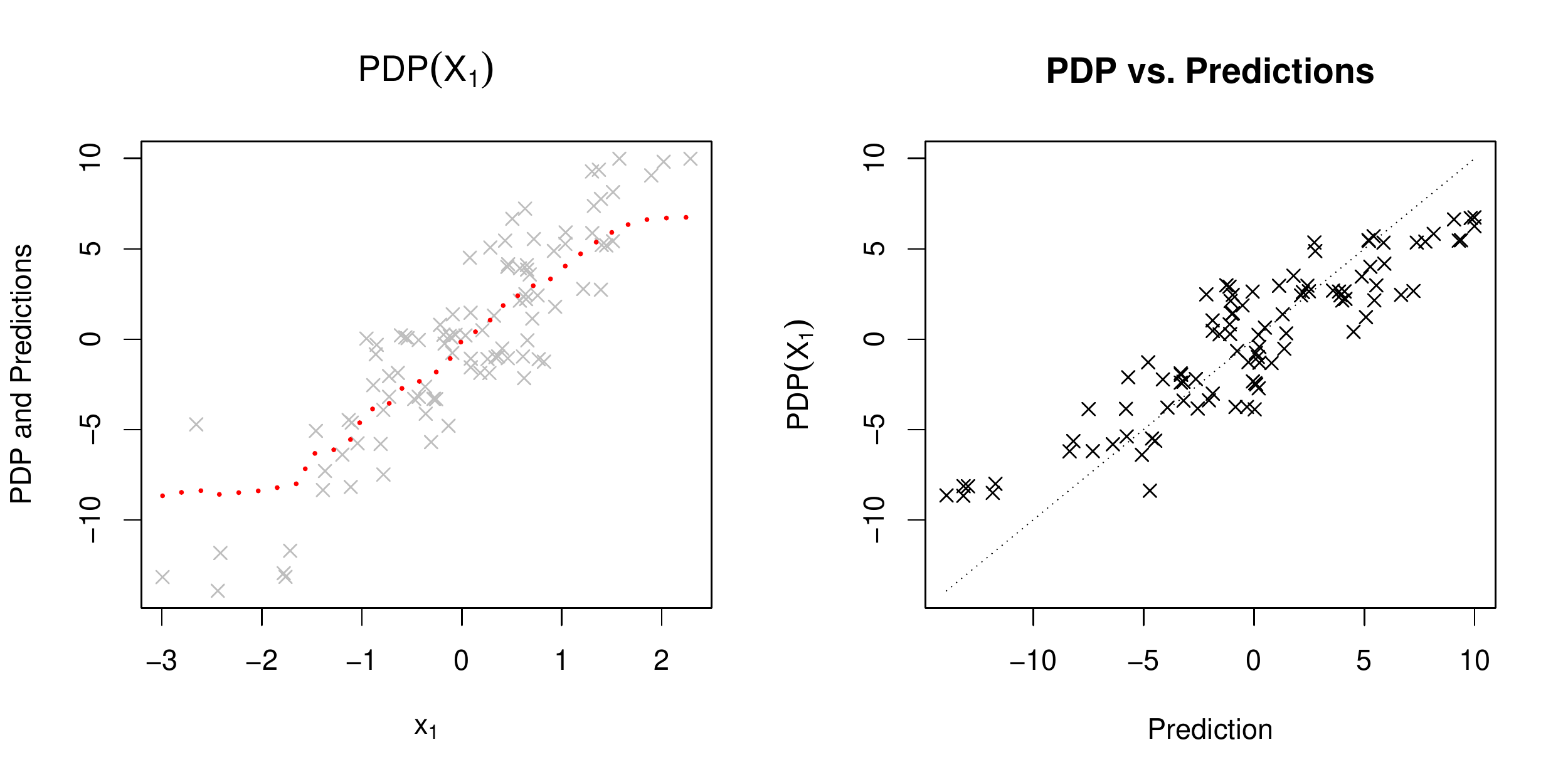}
\end{center}
\caption{PDP for variable $X_1$ (left) and match of partial dependence function $PD(X_1)$ and predicted values $\hat{f}(x)$.}
\end{figure}

In the rest of the paper a measure is defined in order to quantify up to what degree this visualization as given by a PDP is able to explain a model. As an introductory example consider simulated data of two independent random variables $X_1, X_2 \sim N(0,1)$ and a dependent variable $Y$ according to the data generating process:

\begin{equation} \label{eqn:sim1} 
	Y = a X_1 + b X_2 + \epsilon,  
\end{equation}

with $a = 5$, $b = 3$ and a standard normally distributed error term (note that the error term could also be omitted, here). $Y$ depends linearly on $X_1$ and $X_2$. Afterwards a default random forest model \citep[using both variables $X_1$ and $X_2$ and the \texttt{R} package \texttt{randomForest},][]{rf} is computed. Figure 1 (left) shows the corresponding partial dependence plot for variable $X_1$ together with the predictions for all observations. It can be recognized that -- of course -- the PDP does not exactly match the predictions. In Figure 1 (right) the x-axis is changed: here, the predictions of the model $\hat{f}(x_i)$ (x-axis) are plotted against their corresponding values of the partial dependence function $PD_{X_1}(x_{i1})$ (y-axis). The better the PDP would represent the model the closer the points should be to the diagonal. 

\begin{figure}\label{fig:sim2}
\begin{center}
	\includegraphics[width=0.9\textwidth]{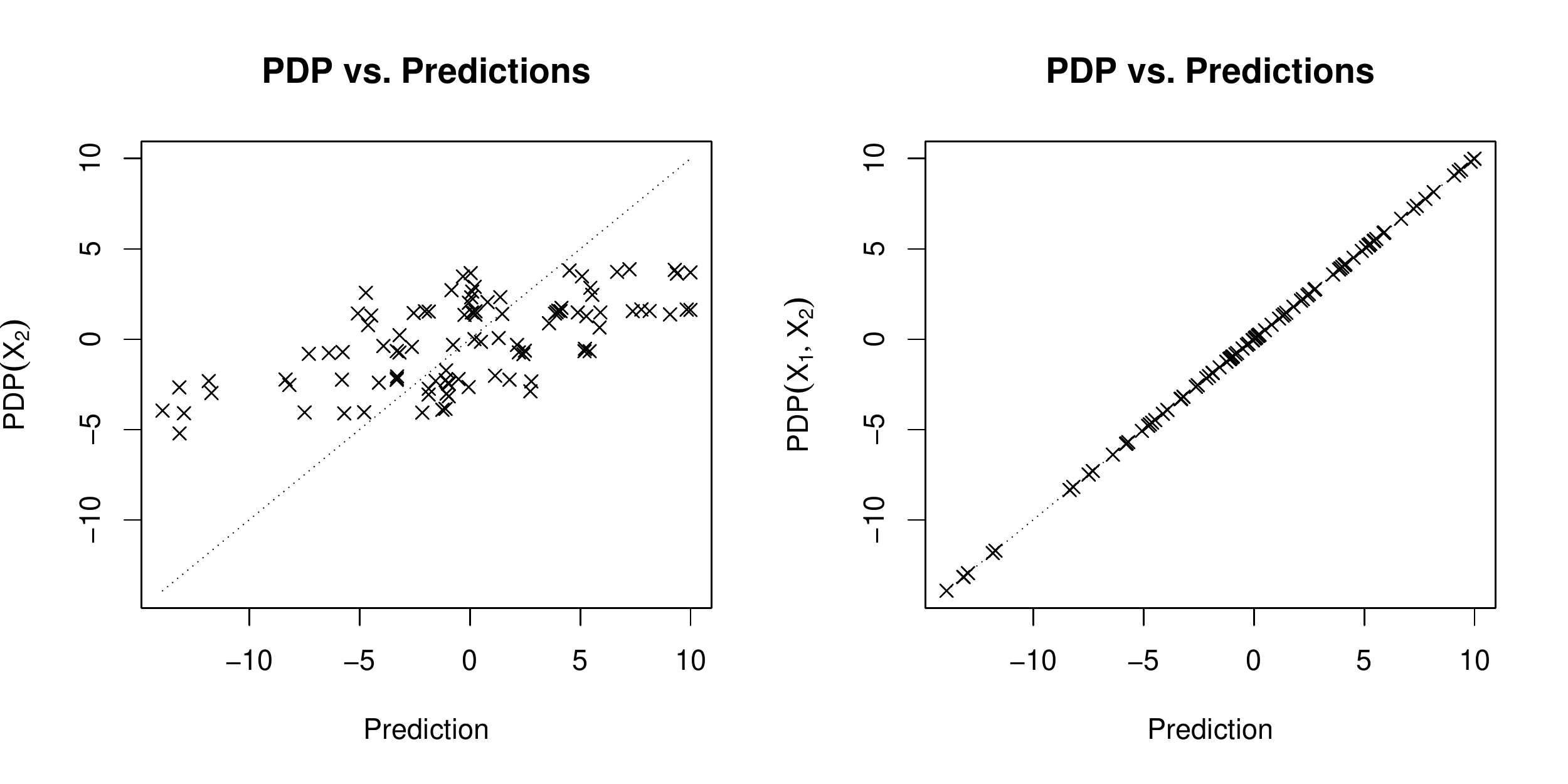}
\end{center}
\caption{Match of partial dependence function $PD(X_2)$ and predicted values for the first example (left). The plot on the right illustrates, that for a 2D-PDP using all input variables $X_s = X$ a perfect match is obtained.}
\end{figure}

A first step towards defining explainability consists in answering the question: \emph{How close is what I see to the true predictions of the model?} 
For this reason, a starting point for further analysis is given by computing the differences between the partial dependence function $PD_s(X_s)$ and the model's predictions. A natural approach to quantifiying these differences is given by computing the expected squared difference:

\begin{equation}\label{asepd}
ASE(PD_s) = \int \left(\hat{f}(X) - \widehat{PD_s}(X) \right)^2  dP(X_s), 
\end{equation}
which can be empirically estimated by:
\begin{equation}\label{asepdemp}
\widehat{ASE}(PD_s) = \frac{1}{n} \sum_{i=1}^n \left(\hat{f}(x_i) - \widehat{PD_s}(x_i) \right)^2. 
\end{equation}

Remarkably the $ASE$ does not calculate the error between model's predictions and the obervations but between the partial dependence function and the model's predictions here.
Further, in order to benchmark the $ASE(PD_s)$ of a partial dependence function it can be compared to the $ASE(PD_{\emptyset})$ of the naive constant average prediction $PD_{\emptyset}$:
\begin{equation}\label{varf}
ASE(PD_{\emptyset}) = \int \left( \hat{f}(x) - PD_{\emptyset} \right)^2  dP(X_s), 
\end{equation}
and its empirical estimate: 
\begin{equation}\label{ssepd}
\widehat{ASE}(PD_{\emptyset}) = \frac{1}{n} \sum_{i=1}^n \left( \hat{f}(x_{i}) - \widehat{PD}_{\emptyset} \right)^2. 
\end{equation}

Finally one can relate both $ASE(PD_s)$ and $ASE(PD_{\emptyset})$ and define {\bf explainability $\Upsilon$} of any black box model $\hat{f}(X)$ by a partial dependence function $PD_s(X)$ by the ratio
\begin{equation}\label{xty}
\Upsilon(PD_s) = 1 - \frac{ASE(PD_s)}{ASE(PD_{\emptyset})} 
\end{equation}
similar to the common $R^2$ goodness of fit statistic: $\Upsilon$ close to 1 means that a model is well represented by a PDP and the smaller it is the less of the model's predictions are explained.

%Note that ICE plots \citep{goldstein15} visualize the variability with regard to the predictions for each single observation in all variables $X_{c}$ which are averaged out by the partial dependence function $PD(X_s)$. 
%In contrast, here the variability with regard to the distribution of $X_s$, i.e. the variables under investigation is considered.   

\section{Examples}\label{examples}
%\vspace*{0.6cm}
%{\bf A first example}:

Starting again with the introductory example from the previous Section.
From data generation the choice of $a > b$ results in a higher variation of $Y$ with regard to $X_1$. Accordingly, it can be expected that $PD(X_1)$ is closer to the model's predictions than the $PD(X_2)$ (cf. Figure~2, left) and thus has a higher explainability.
Computing both explainabilities confirms this: $\Upsilon(PD_{X_1}) = 0.786 > 0.356 = \Upsilon(PD_{X_2})$. For a two dimensional PDP the partial dependence function corresponds to the true predictions resulting in an explainability of 1, i.e. the model is perfectly explained by the partial dependence curve (Figure~2, right).

\begin{figure}\label{fig:boston}
\begin{center}
	\includegraphics[width=0.9\textwidth]{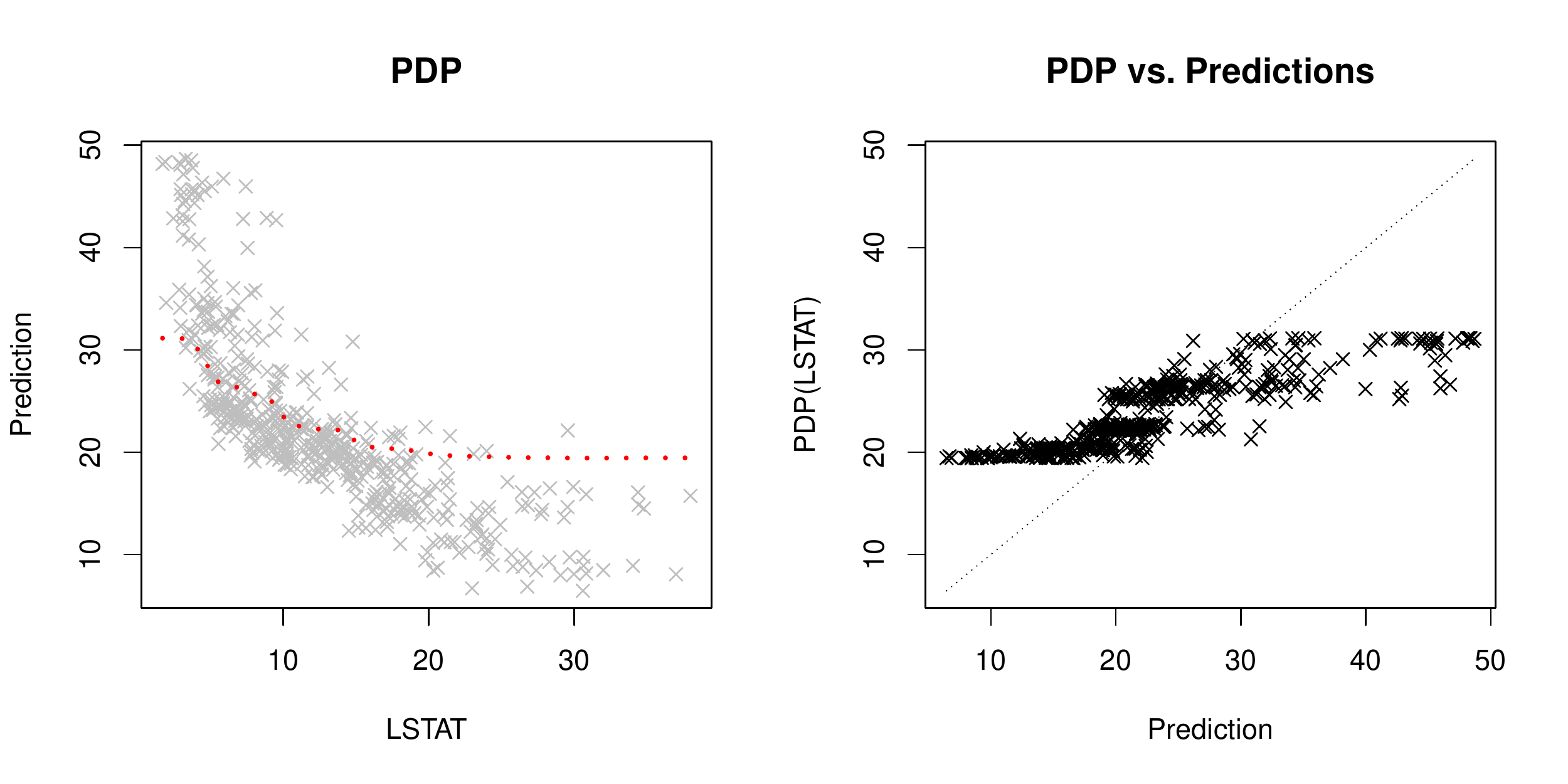}
\end{center}
\caption{Most explainable PDP for a random forest model on the boston housing data (left) as well as match of preditions and PDP (right).}
\end{figure}

%\vspace*{0.5cm}
%{\bf Example 2: Boston housing data}:

\begin{table}[ht] \label{tab1:boston1d}
\centering
\begin{tabular}{cc|cc}
  \hline
Variable & $\Upsilon$ & Variable & $\Upsilon$ \\ 
  \hline
  lstat & 0.512 & age & 0.018 \\ 
  rm & 0.410 & b & 0.012 \\ 
  lon & 0.085 & chas & 0.004 \\ 
  nox & 0.056 & zn & 0.002 \\ 
  ptratio & 0.056 & lat & 0.001 \\ 
  indus & 0.046 & rad & -0.002 \\ 
  tax & 0.030 & dis & -0.004 \\ 
  crim & 0.025 &  &  \\  
   \hline
\end{tabular}
\vspace*{0.2cm}
\caption{Explainability of 1D PDPs for a random forest model of the Boston housing data based on different variables.}
\end{table}

As a $2^{nd}$ example the popular Boston housing real world data set~\citep{uci} is used which has also been done by other authors \citep[cf. e.g.][]{pdp} in order to illustrate partial dependence plots. Again, a default random forest model has been built as in the example before. Figure~3 (left) shows the PDP for variable \texttt{LSTAT}. The corresponding explainabilities identify these PDPs to be the two most useful ones (cf. Table~\ref{tab1:boston1d}). 

Nonetheless, from the explainabilities of all single variable's partial dependence functions it is also obvious that considering single PDPs alone is not sufficient to understand the behaviour of the model in this case. Taking a closer look at the partial dependence function vs. the predicted values on the data set (Figure~3, left) shows further that e.g. for large values of the variable \texttt{LSTAT} the partial dependence function (dotted red line) appears to systematically overestimate the predictions for this example. %This means, that large values of \texttt{LSTAT} tend to fall together with values in (interact with) other variables which decrease predictions relative to the distribution of the data within these variables ($X_c$)~\citep[cf. also][]{apley16}.  

Taking into account for the explainability $\Upsilon$ provides us with the information of how strong the true predictions deviate from what we do see in the partial dependence plot. Comparison of Figure~3 (right) and Figure~4, (right) illustrates that the two dimensional PDP of the two most explainable variables \texttt{LSTAT} and \texttt{RM} is much more explainable in this case ($\Upsilon = 0.759$). A coloured scatterplot can be used in order to visualize the two dimensional PDP together with the distribution of the observations in both variables (Figure~4, left) as well as the gap between the partial dependence function and the predicted values by the model (Figure~4, center) where in both plots blue represents low (/negative) and red represents high (/positive) values.
  
\begin{figure}\label{fig:boston2d}
\begin{center}
	\includegraphics[width=0.95\textwidth]{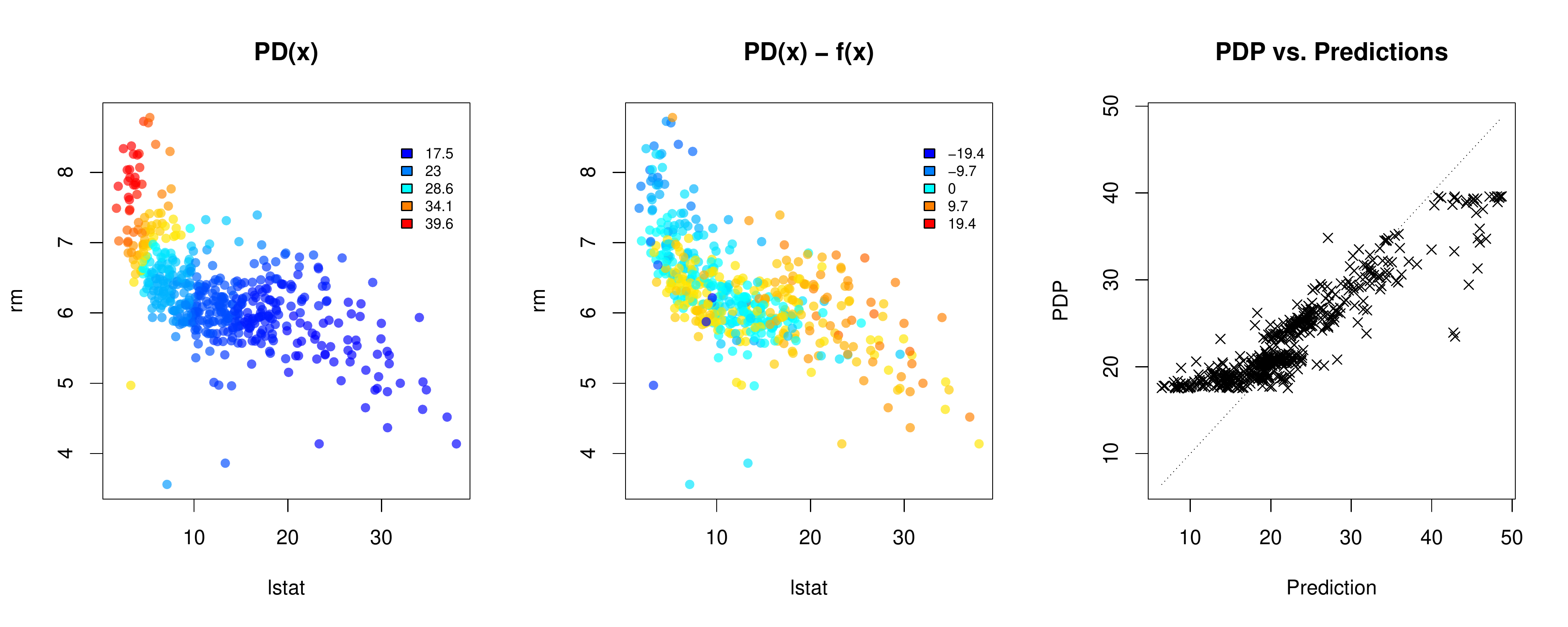}
\end{center}
\caption{Two dimensional PDP for the variables variables \texttt{LSTAT} and \texttt{RM} of a random forest on the Boston housing data (left) and the corresponding match of PDP and preditions(right).}
\end{figure}

Generally, partial dependence functions are not resticted to 1D or 2D thus one can analyze how much the partial dependence function gets closer to the model if we include additional variables. Table~2 shows the results of a forward variable selection using $\Upsilon$ as selection criterion for stepwise inclusion of variables in $X_s$:

\begin{table}[ht] \label{tab1:boston1d}
\centering
\begin{tabular}{rlr|rlr}
  \hline
  Step & Variable & Ups & Step & Variable & Ups \\ 
  \hline
  1 & lstat & 0.512 &   9 & tax & 0.986 \\ 
  2 & rm & 0.759 &  10 & age & 0.993 \\ 
  3 & lon & 0.805 &  11 & b & 0.997 \\ 
  4 & nox & 0.847 &  12 & lat & 0.999 \\ 
  5 & crim & 0.894 &  13 & rad & 0.999 \\ 
  6 & dis & 0.931 &  14 & chas & 1.000 \\ 
  7 & ptratio & 0.958 &  15 & zn & 1.000 \\ 
  8 & indus & 0.974 &  &  &  \\ 
   \hline
\end{tabular}
\vspace*{0.2cm}
\caption{Results of forward variable selection for $X_s$ according to maximise explainability $\Upsilon$ for the Boston housing data.}
\end{table}

It can be seen that with as few as six variables an explainability of $\Upsilon > 0.93$ is obtained. Nonetheless, there is still need for collection of experiences what level of $\Upsilon$ could be considered as a sufficient explanation of a model. 
Furthermore, although it is principally possible to compute partial dependence for vectors $X_s$ of any dimension its visualization is restricted to $dim(X_s) \le 2$.  
%which is no limit for the proposed visualization of the match between $PD$ and $\hat{f}$ as the one in Figure~4 (right). 
As an attempt to consider more than two variables at once, a scatterplot matrix of two-dimensional PDPs can be computed. Figure~5 shows such a scatterplot matrix for the first four variables of the Boston housing data according to $\Upsilon$-based variable selection: It can be easily recognized from the plot that a high number of rooms \texttt{rm} as well as a low percentage of habitants with lower status \texttt{lstat} are most important for prediction of high house prices as well as a cooccurence of both. But this visualization of course still fails to visualize nonlinear dependencies of higher order which are potentially also identified by the model $\hat{f}(x)$ and it remains an open topic for future resarch activities to develop methodologies to understand high order interactions of variables within models, e.g. based on the works of \citet{britton19} and \citet{gosiewska19}. 
In contrast, some authors suggest restricting to interpretable models \citep{rudin19} which often \citep[but not always, cf. e.g.][]{buecker19} trades off with predictive power. In general the benefit of using a rather complex models should be analyzed for each situation separately. \citet{molnar19} suggest a strategy for simultaneously optimizing a compromise between accuracy and interpretability.

\begin{figure}\label{fig:matrix}
\begin{center}
	\includegraphics[width=0.75\textwidth]{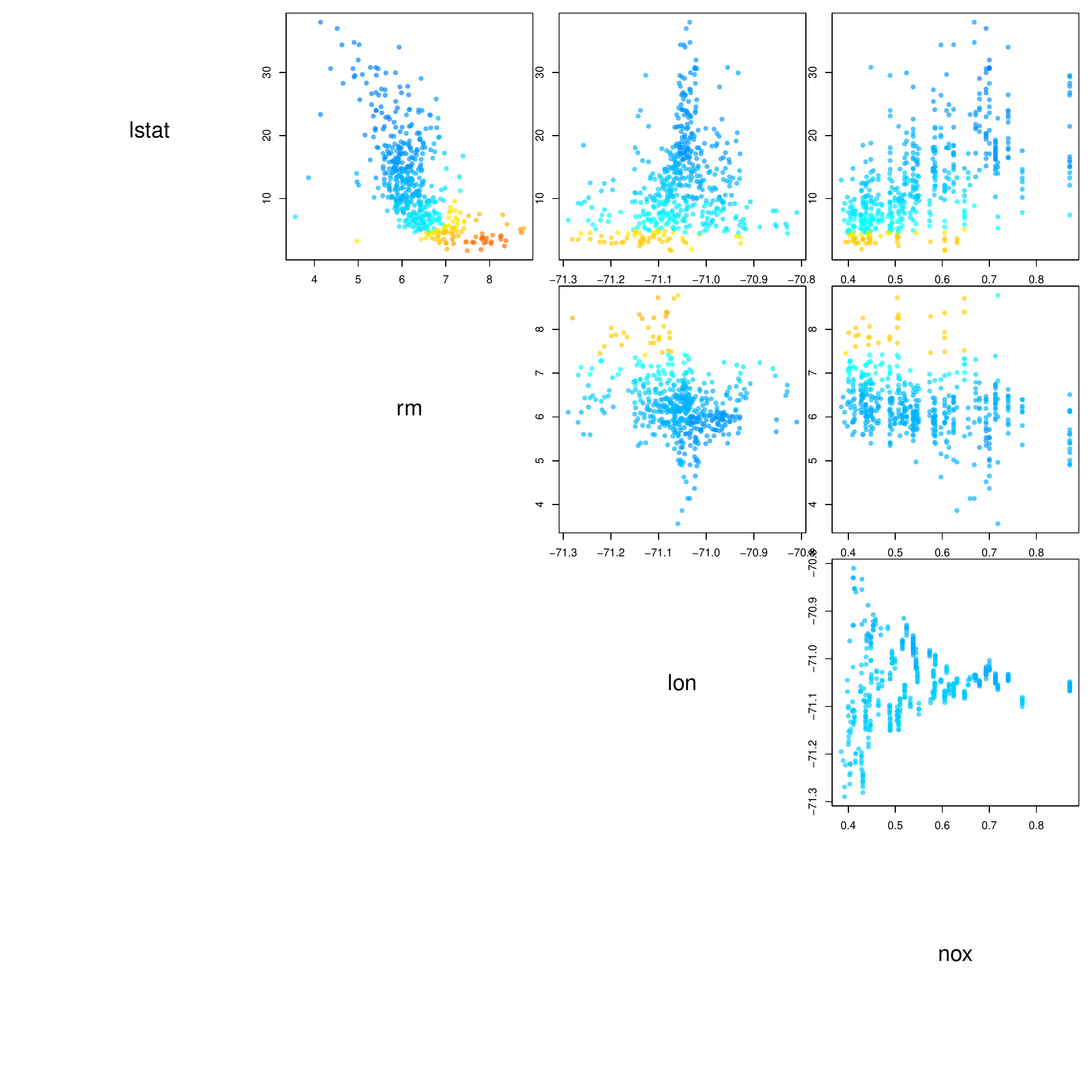}
\end{center}
\caption{Scatterplot matrix of 2D PDPs for the first four variables according to $\Upsilon$-based variable selection.}
\end{figure}

\section{Summary}\label{summary}
Partial dependence plots as one of the most common tools to explain feature effects of black box machine learning models are investigated with regard to the extent that they are able to explain a model's predictions. %As a restiction their visualization is only possible for low order dependencies. 

Using differences between the predictions of the model and their corresponding values of a partial dependence function a framework has been developed to quantify how well a PDP is able to explain the underlying model. The result in terms of the measure of \emph{explainability} $\Upsilon$ allows to assess whether the explanation of a black box model may be sufficient or not. 

As a graphical approach to assess explainability the match between the partial dependence function and the model's output as a scatterplot of the data in the $(\hat{f}(x), PD(x))$ plane is proposed. For two-dimensional PDPs the differences between both functions can be visually localized in a coloured scatterplot of $(PD(x) - \hat{f}(x))$. 

Two simple examples have been presented in order to illustrate the concept of explainability. It can be seen that looking at PDPs is not necessarily sufficient to understand a model's behaviour. As an open issue it has to be noted that although PDP visualizations are restricted to dimensions lower or equal than two the models in general use more than two variables. Of course, analysts are able to look at several PDPs at the same time, e.g. using scatterplot matrices of partial dependence plots but up to our knowledge no literature is available howfar humans are able to combine information of more than two PDPs in order to get a clearer picture of a model's behaviour which remains as a topic of future research.

%\paragraph{Herangehensweise}

%\pagebreak

\bibliographystyle{spbasic}
\bibliography{szepannek_explainability_V2}

\end{document}